\documentclass[letterpaper, 10 pt, conference]{ieeeconf}  

\IEEEoverridecommandlockouts                              

\usepackage[left=54pt,right=54pt,top=54pt,bottom=54pt]{geometry}

\usepackage{graphicx} 
\usepackage{epsfig} 
\usepackage{soul}
\usepackage{comment}
\let\labelindent\relax
\usepackage{enumitem}

\usepackage{appendix}
\usepackage[caption=false]{subfig}
\usepackage{multicol}
\usepackage{cite}
\usepackage{color}
\definecolor{mygray}{gray}{0.6}
\newcommand{\DD}[1]{{\color{black}#1}}
\newcommand{\SV}[1]{{\color{black}#1}}

\usepackage{cuted}
\usepackage{capt-of}
\usepackage{gensymb}
\usepackage{hyperref}
\title{\LARGE \bf
Evaluation of Cross-View Matching to Improve Ground Vehicle
Localization with Aerial Perception
}

\author{Deeksha Dixit, Surabhi Verma, Pratap Tokekar
\thanks{This work is supported by the Office of Naval Research under Grant No. N000141812829.}
\thanks{We thank Nathaniel M Glaser and Zsolt Kira from Georgia Institute of Technology for help with the AirSim setup.}
\thanks{D. Dixit and P. Tokekar were with the Department of Electrical
and Computer Engineering, Virginia Tech, USA when part of the work was completed. All authors are currently with the University of Maryland at College Park, USA. 
{\tt\small \{deeksha, sverma96, tokekar\}@umd.edu}).}
}

\begin{document}
\maketitle


\begin{strip}\centering
\vspace{-90pt}
\includegraphics[width=1\textwidth]{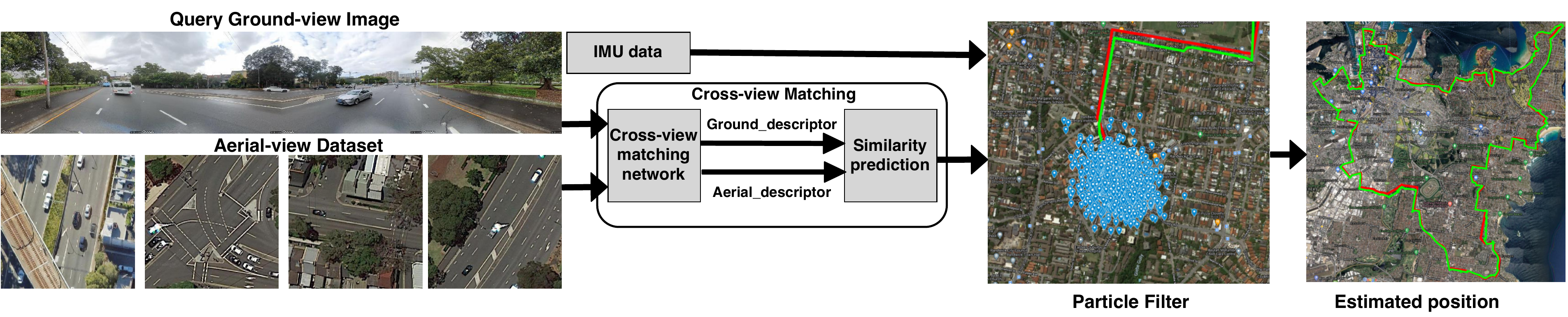}
\captionof{figure}{
Overview of the cross-view localization pipeline. A query ground view image and an aerial view dataset are inputs to the system. A cross-view matching network, CVM-Net-I~\cite{hu2018cvm}, is used to generate viewpoint invariant descriptors. This information is fed to a particle filter which samples locations at every timestep to give position estimates. The ground truth (green) and estimated trajectory by the particle filter (red) is highlighted on the right.
}
\label{fig:sytem_architecture}
\end{strip}
\begin{abstract}
Cross-view matching refers to the problem of finding the closest match \DD{of} a given query ground view image to one from a database of aerial images. If the aerial images are geotagged, then the closest matching aerial image can be used to localize the query ground view image. Due to the recent success of deep learning methods, several cross-view matching techniques have been proposed. These approaches perform well for the matching of isolated query images. However, their evaluation over a trajectory is limited. In this paper, we evaluate cross-view matching for the task of localizing a ground vehicle over a longer trajectory. \DD{We treat these cross-view matches as sensor measurements that are fused using a particle filter}. We evaluate the performance of this method using a city-wide dataset collected in a photorealistic simulation by varying \DD{four} parameters: height of aerial images, the pitch of the aerial camera mount, FOV of the ground camera, and \DD{ \SV{the} methodology of fusing cross-view measurements in the particle filter}. \DD{We also report the results obtained using our pipeline on a real-world dataset collected using Google Street View and satellite view APIs.

}


\end{abstract}

\section{INTRODUCTION}
Consider a ground vehicle that is navigating in an environment. Localizing this vehicle in a global frame of reference (\emph{geolocalization}) is critical for efficient planning. Geolocalization can be achieved by using a GPS onboard the vehicle. However, GPS can be noisy and unavailable at times, especially when operating in urban environments with tall buildings \DD{or} canopies~\cite{whyatt2008noisy}. In such cases, the localization can be improved by using onboard vehicle perception (e.g., stereo, inertial sensors, and LIDAR). In this paper, we study a technique to complement onboard perception with cross-view matching for localization in a global frame \DD{over a trajectory}.


Cross-view matching is the problem of finding an aerial image in a database of aerial images that is the closest match to a given ground view query image~\cite{Castaldo_2015_ICCV_Workshops}~\cite{zhu2020leveraging}. In order to match images taken from vastly different viewpoints, learning is required. This problem has applications in situations such as identifying the location where a photo was captured ~\cite{workman2015wide} and guiding ground based navigation. Specifically, cross-view matching can be used for cross-view localization if the aerial images are georeferenced. The aerial images can be satellite images or can be images that were collected by an aerial vehicle from a lower altitude. Every aerial image has latitude and longitude information of where the image was taken from. By matching a ground view to an aerial database we can predict the location of the query ground image. 

\DD{Prior work has shown the potential for cross-view matching in order to localize ground vehicles using satellite imagery ~\cite{viswanathan2014vision,shan2014accurate,lin2013cross}. More recently, deep learning has achieved great success in cross-view image retrieval}. \DD{Several of these works \cite{lin2015learning,hu2018cvm, liu2019lending} use a modified two-branch Siamese neural network~\cite{koch2015siamese} that is trained on a database of paired aerial and ground images to generate image descriptors which are robust to large viewpoint changes. These networks learn to predict embeddings which are closer in feature space for a true aerial-ground image pair and far off for a non-matching pair. Consequently, the distance between these image descriptors or embeddings can be used as a similarity measure between an aerial and a ground image. This metric can be used during test time to compute the similarity of a ground view query image with each image in the database of aerial images. The image with the highest similarity score or the set of $k$ images with the top $k$ highest scores, are considered as the closest matching aerial images.} 

\DD{The output of such networks can be thought of as a noisy position observation of the ground vehicle. Over time, these noisy position observations can be fused using an estimator such as a particle filter~\cite{Fox2001ParticleFF}. In this paper, we perform a thorough evaluation of such a pipeline (Figure~\ref{fig:sytem_architecture}) which deals with drastically different aerial and ground viewpoints.} 

\DD{We use a recently proposed architecture for cross-view matching, called CVM-Net~\cite{hu2018cvm}, to generate descriptors for aerial and ground images. Project Autovision \cite{DBLP:journals/corr/abs-1809-05477} and extended work on CVM-Net \cite{hu2020image} showed one way of integrating the output from CVM-Net in a particle filter to localize a ground vehicle. While these results are promising, their evaluation was restricted to one specific dataset, and one specific method of integrating cross-view matching and particle filtering. In general, there are several design choices one has to make that may affect the localization accuracy. }

Our goal in this paper is to, therefore, conduct an exhaustive empirical analysis of the following four design choices:
\begin{enumerate}
    \item \DD{The method with which the output of the \DD{Cross-view matching network} can be integrated as an observation in the particle filter. There exists a choice of using only the top $k$ matches~\cite{Doan2019VisualLU}, or all the similarity scores~\cite{DBLP:journals/corr/abs-1809-05477} between the query ground image and \emph{all} the aerial images in the database. We conclude that using all the similarity scores works better.} 
    \item The height at which the aerial images are obtained. Prior evaluation was limited to images taken from one altitude. \DD{We find that the retrieval accuracy is better for aerial images taken from a lower altitude but the localization performance remains unaffected.} 
    \item The field of view (FOV) of the ground view image. Prior evaluation only used a panoramic ground view image (which can see more local information). \DD{Our study indicates that both, retrieval and localization performance, is consistent across FOVs.} 
    \item The camera pitch of the aerial images. Prior work only used top-down images. \DD{Our results show that images taken with a pitch of -50$\degree$ (look-ahead) give a significant boost in both the retrieval and localization performance over top-down images.} 
\end{enumerate}

\DD{We evaluate cross-view matching} in the context of these design choices. We \DD{collect} a large dataset of ground and geotagged aerial images in AirSim~\cite{DBLP:journals/corr/ShahDLK17}, a photo-realistic simulator. Using this dataset, we conduct several numerical experiments to study the performance of \DD{cross-view matching} \SV{and} localization. Our results can be helpful for a practitioner who is interested in using \DD{cross-view matching} for supporting onboard localization and eliminating the guesswork that is typically involved when making such design choices.

\DD{In addition to our experimental results, the dataset we have created in AirSim\footnote{\url{https://drive.google.com/drive/folders/1ztD8iAa3_PGpFv_up8FizlqkqdfUngmT?usp=sharing}} is another contribution of this paper. Existing datasets such as CVUSA \cite{workman2015localize}, CVACT \cite{liu2019lending} and the recently introduced University1652 \cite{zheng2020university} are standard cross-view matching datasets. While they are useful for evaluating cross-view matching, they do not contain a continuous trajectory of ground-view images that one would need in the work we present. Furthermore, these datasets do not include all scenarios that can affect the performance of a cross-view localization pipeline. To remedy this, we collected our own dataset on AirSim to model a variety of cross-view data collection settings. We extended \SV{a part of} this evaluation on real-world data by using a dataset from Google Street View and $45\degree$ imagery from Google Maps API.}

The rest of this paper is organized as follows. We start with the related work in Section~\ref{related work}. Then, we describe the overall architecture of the system that integrates \DD{cross-view matching} and particle filtering, in Section~\ref{sec:methodology}. The experimental setup and results are discussed in Section~\ref{experiments}. Finally, we conclude with a discussion of the \DD{future} work in Section~\ref{sec:conclusion}.




\section{RELATED WORK}
\label{related work}
Research on identifying the location of an image has been studied mostly as an image retrieval problem~\cite{tian2017cross}. Here, one image from a database must be retrieved which closely matches the query image.
Hays et al.~\cite{hays2008im2gps} proposed a technique, termed IM2GPS, which uses geotagged photos from Flickr and then models the query image as a probabilistic distribution over the entire world. In \cite{agarwal2009building, irschara2009structure,li2010location} the authors developed a 3D model for localization. In these approaches, the query and database images to be searched have been collected from similar views.  

\DD{The problem of visual place recognition and localization has been extensively studied for typical scenarios where viewpoints of reference and query images are similar, such as views obtained by the front and rear facing cameras on the same vehicle. These are not as extreme as ground and aerial views. Sarlin et al.~\cite{sarlin2019coarse} solve the 6 DoF localization by proposing HFNet that uses a monololithic CNN which predicts local features and global descriptors in a hierarchical approach. \SV{Cipolla et al. \cite{kendall2015posenet} use transfer learning to use a CNN, trained on \textit{ImageNet}~\cite{imagenet_cvpr09}, as a pose regressor in a global frame. The authors in \cite{mapnet} propose an extension of this work, where, additional relative pose along with Visual Odometry and GPS data is used for better pose convergence.} 

There have also been attempts to use scene semantic information to learn descriptors that are robust to viewpoint and appearance changes \cite{doi:10.1177/0278364919839761}. Gawel et al.~\cite{gawel2018x} perform global localization of a robot over a longer trajectory by transforming a semantically segmented image into query graphs. 
Evaluations are performed using a synthetic dataset collected from SYNTHIA~\cite{synthia} as well as from Google street dataset with front-facing and rear-facing ground view images. They report a localization accuracy of 30\textit{m} on a dataset collected in AirSim. However, perfect semantic segmentation might not always be available in real-world scenarios.}

\DD{Combining image retrieval with particle filter has been explored in the literature before. Wolf et al.~\cite{wolf2005robust} and Menegatti et al.~\cite{menegatti2004image} use a combination of image retrieval and sample-based Monte Carlo localization technique to localize a robot in an indoor environment. More recently, Xu et al.~\cite{xu2019robust} proposed a more robust indoor localization pipeline using CNN-based image retrieval from the same viewpoint. Doan et al.~\cite{Doan2019VisualLU} address the problem of 6 DoF localization under appearance changes in an outdoor environment. 
Project Autovision~\cite{DBLP:journals/corr/abs-1809-05477} and Hu et al.~\cite{hu2020image} use CVM-Net to generate the image embedding. The resulting pipeline in~\cite{DBLP:journals/corr/abs-1809-05477} yields a localization error of 9.92\textit{m} in an urban environment and 9.29\textit{m} in a rural environment over a route of 5\textit{km}. Furthermore,~\cite{hu2020image} achieves an error of 16.39\textit{m} in urban and 20.33\textit{m} in a rural environment on a trajectory of 5\textit{km} and 3\textit{km}, respectively in Singapore.}

\DD{The second approach to geolocalization is the cross-view approach which is the focus of our work here. Cross-view, as defined earlier, involves taking images from different viewpoints to solve the problem of geolocalization. Zhang et al.~\cite{4155707} used SIFT-based image matching and use the average of the top three images to obtain the geotags for the query image. In \cite{tian2017cross}, the authors used Faster R-CNN~\cite{DBLP:journals/corr/RenHG015} to identify buildings in the query image and the testing set.}

\DD{Hu et al. use a novel technique of using a generalized VLAD~\cite{arandjelovic2013all} layer called NetVLAD ~\cite{Arandjelovi2016NetVLADCA} on top of CNN to extract view-point invariant image descriptors. 
The resulting system yields 37\% accuracy using top 1 and 91.4\% accuracy using the top 80 image retrievals from the aerial dataset. Liu et al.~\cite{liu2019lending} further improve localization performance by using additional orientation information. Regmi et al.~\cite{regmi2019bridging} use Generative Adversarial Networks (GANs) through a joint-feature-fusion network, to get a more robust representation. Shi et al. ~\cite{shi2019spatial} use polar transformation on the aerial image and spatial-attention mechanism to get further boost in the retrieval performance. We chose CVM-Net to generate embedding for our localization pipeline. However, all the aforementioned pipelines generate the image descriptors similar to CVM-Net for retrieval. We hypothesize that the results from this paper should be valid even for all these recent cross-view matching networks.}

\DD{Our focus in this paper is not on designing a new cross-view matching algorithm but instead on rigorously evaluating how cross-view matching performs when it comes to localizing a ground vehicle. Prior work only evaluate for a specific design choice and in very limited settings. We thoroughly evaluate the space of design choices to find  the effect of cross-view matching on estimating a trajectory of the vehicle.}




 





\section{Cross-View Matching Based Localization Framework}
\label{sec:methodology}
\newcommand{\norm}[1]{\| #1 \|}
In the context of this paper, geolocation refers to the 2D position $X_{t}=(x,y)$ of an agent with respect to some global frame of reference. Our framework for tracking a ground vehicle using cross-view matching and particle filter is shown in Figure~\ref{fig:sytem_architecture}. We assume that the ground vehicle has an initial estimate of its position in the global frame, and is equipped with an Inertial Measurement Unit (IMU) and a forward-facing RGB camera. We also assume that there is an aerial view dataset collected by a drone (or some low altitude aircraft) with a downward-facing camera. Each image in the aerial view dataset is a top-down view of the environment with associated geolocation information. 

As shown in Figure~\ref{fig:sytem_architecture}, a ground view query image from the forward-facing camera is given as an input to the system along with the geotagged aerial view dataset. The output produced by the localization system is the estimated position of the ground vehicle. Our proposed localization approach is a two-step procedure. First, we convert the ground view query image and the aerial view image into view-point invariant image descriptors for assessing the similarity between them. This is achieved using a Siamese based neural-network CVM-Net-I.
Secondly, we use a particle filter to fuse observations over a trajectory. 

Particle filters consist of three primary steps: sampling, prediction, and update. First, $M$ particles are sampled from the prior distribution.  In the prediction step, the IMU information from the ground robot is used to propagate the particles at each time step. Our system uses the similarity information from the first step to update the weights of the particles in accordance with their posterior likelihood. Then, the weights of these particles are normalized such that the sum of all particle weights adds up to 1. Finally, a new set of particles is sampled from this weighted distribution of particles \DD{using stochastic universal sampling~\cite{whitley1994genetic}}. The estimated ground robot position is a weighted mean of the resampled particles at each time step.

We propose two strategies for updating the weights of the particles. We call these techniques Prediction-based Particle Filtering (PPF) and Compare-All Particle Filtering (CAPF).  

\paragraph*{Prediction-based Particle Filtering (PPF)}
\begin{figure}
      \centering
      \includegraphics[width=\columnwidth]{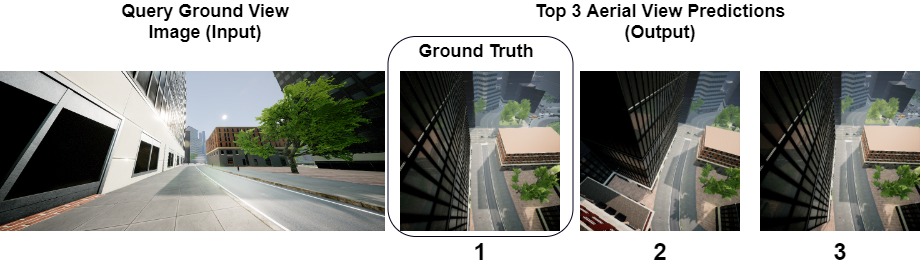}
      \caption{Illustration of image retrieval. For the ground view query image shown on the left top 3 retrievals from the aerial database are visualized on the right. The top 1 retrieval in this case is also the ground truth.}
      \label{fig:predictions}
 \end{figure}

\DD{Cross-view matching networks} solve a retrieval problem. The primary aim of retrieval is to find the $k$-nearest-neighbors for a ground view query image in an aerial view dataset. In PPF, the posterior distribution is updated using the top $k$-nearest-neighbors from the aerial view dataset, for \DD{a given query image. The value of $k$ is predetermined (We evaluate the effect of choosing $k$).}  Figure~\ref{fig:predictions} indicates the top 3 retrievals for the ground view query image shown on the left.

\DD{Our PPF pipeline is based on the one given in \cite{Doan2019VisualLU}}. \DD{Primarily, we adopt the preprocessing step of mean shift clustering used in this work.} First, we pass the query image and all the aerial view images through the CVM-Net to convert them to image descriptors. Each descriptor has an individual dimension of $1\times4096$. Then at every time step, we retrieve the top $k$-nearest-neighbours for the ground view for that time step using the \DD{L2 distance between the query image descriptor and all the aerial view descriptors. Next, we apply mean shift clustering algorithm~\cite{carreira2015review} on the poses of all the retrieved aerial images. The mean of poses of all the images in the largest cluster is used as the noisy measurement $z_t$ for that time step. Finally the weight of each particle is updated using: }
$w_{t}^{[i]}~\propto~e^{-\frac{1}{2}\left(z_{t}-p_{t}^{[i]}\right)^{T} \Sigma_{o}^{-1}\left(z_{t}-p_{t}^{[i]}\right)}$.
Here, $w_{t}^{[i]}$ is the weight assigned to particle $i$ at time $t$, $p_{t}=(x,y)$ is the position at time $t$, and $\Sigma_{o}^{-1}$ is the noise covariance matrix which models the noise in the measurements.
 
\paragraph*{Compare-All Particle Filtering (CAPF)}
The CAPF methodology is adapted from project Autovision~\cite{DBLP:journals/corr/abs-1809-05477}.
\DD{After getting the image descriptors through CVM-Net, at each time step $t$, we first find an aerial image which has the position $(x,y)$ closest to each particle $i$ and we assign the image descriptor of this image to the particle and call it $I_{t}^{[i]}$. The weight assigned to each particle is inversely proportional to the L2 norm between the image descriptors of the ground view query image $I_{t}^{q}$ at time $t$ and the image descriptor assigned to the particle $I_{t}^{[i]}$ as shown in equation and the aerial view image descriptor with a geotag closest to the particle position:
$w_{t}^{[i]}~\propto~\norm{I_{t}^{q}-I_{t}^{[i]}}^{-1}_2$}.
 
 
\DD{The only difference in these two methodologies is how the image descriptors from the cross-view matching network are used to update the weights. The resampling strategy and how the final position is estimated is the same in both pipelines.}

In the next section, we empirically compare these two methods as well as the study of several other design issues.


\section{Empirical Evaluation and Results}
\label{experiments}
\begin{figure}
      \centering
      \includegraphics[width=0.6\columnwidth]{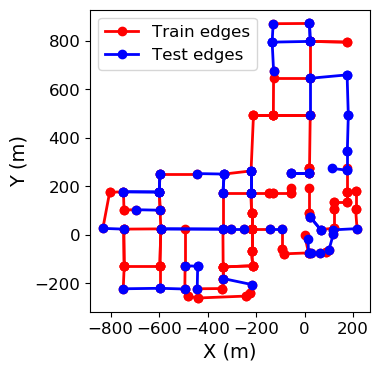}
      \caption{Plot of the trajectories traversed by the aerial and ground robot. The trajectories used for training and testing are marked with red and blue colour.}
      \label{fig:city_graph}
\end{figure}
\subsection{Dataset}
\begin{figure*}
      \centering
      \includegraphics[width=\linewidth]{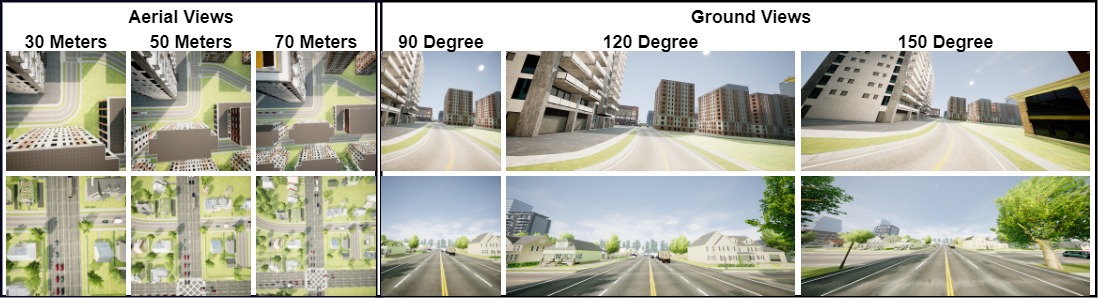}
      \caption{A subset of images from the AirSim dataset. Images in the same row come from the same scene. Images on the left are the aerial views captured from the altitude indicated. Images on the right are ground view for the same scene captured using different FOV.}
      \label{fig:view_grid}
\end{figure*}
For training and evaluation, we generated our dataset using the photorealistic simulation API in AirSim which is a simulator for drones and cars built on the Unreal Engine. It provides the functionality of spawning multiple agents in the environment and provides full control over their movements. It also makes it possible to model different weather and temporal conditions. Our current setup makes use of the pre-compiled binaries from the City Environment in AirSim which is a large environment with moving vehicles and pedestrians. Figure~\ref{fig:city_graph} shows a plot of all the training and test data collection trajectories within the city.  

Our dataset consists of images collected from 5 different altitudes and 3 different FOVs for a single scene. In total, the dataset consists of $5\times3\times2546$ pairs of aerial and ground view images spanning an area of approximately 1544.75\textit{m} diagonally over the trajectories shown in Figure~\ref{fig:city_graph}. We also collected the same number of images for two different pitch values for the downward-facing camera on the UAV flying at an altitude of 50\textit{m} \DD{as shown in Figure~\ref{fig:pitch}}. 

For any one trial, we use 1679 images for training and 867 images for testing. The dataset was collected by flying drones with a downward-facing camera\footnote{All the altitudes mentioned in the paper are reported with respect to a base height of 10\textit{m} at which we flew the drone. To get the absolute value of the altitude in AirSim an offset of +10\textit{m} should be added to all the altitude values reported.} at two  different altitudes on the same trajectory and then fetching the images from these vehicles along with the positional information. One of the drones was flown at a lower altitude of 1\textit{m} to simulate the movement of a ground vehicle. Nonetheless, the ground, and aerial images are not perfectly synced; the paired positions have an inherent noise of $4.58\pm2.44$\textit{m} due to AirSim limitations. Therefore, even in the case of 100\% accurate retrieval, there will be a localization error of this amount which is expected in practice. 

The sample images from the AirSim dataset are shown in Figure~\ref{fig:view_grid}.
When evaluating geolocalization, we use the complete dataset (which includes training and test images since it represents a Complete trajectory through the city) and one that includes only the test dataset (Figure~\ref{fig:city_graph}).


\subsection{Experiments}

In this section, we present the experiments conducted to evaluate the performance of the particle filter localization across system-level design choices and various data collection settings. The system-level parameters include altitude of aerial images, FOV of ground images, and pitch of the aerial view camera. For the evaluation of design choices, we compare the PPF and CAPF methodology mentioned in Section~\ref{sec:methodology}. 

\subsubsection{PPF vs. CAPF}
\begin{figure}
    \centering
     \subfloat[Plot of ground truth, dead reckoning and particle filter trajectories]{{\includegraphics[width=0.9\columnwidth]{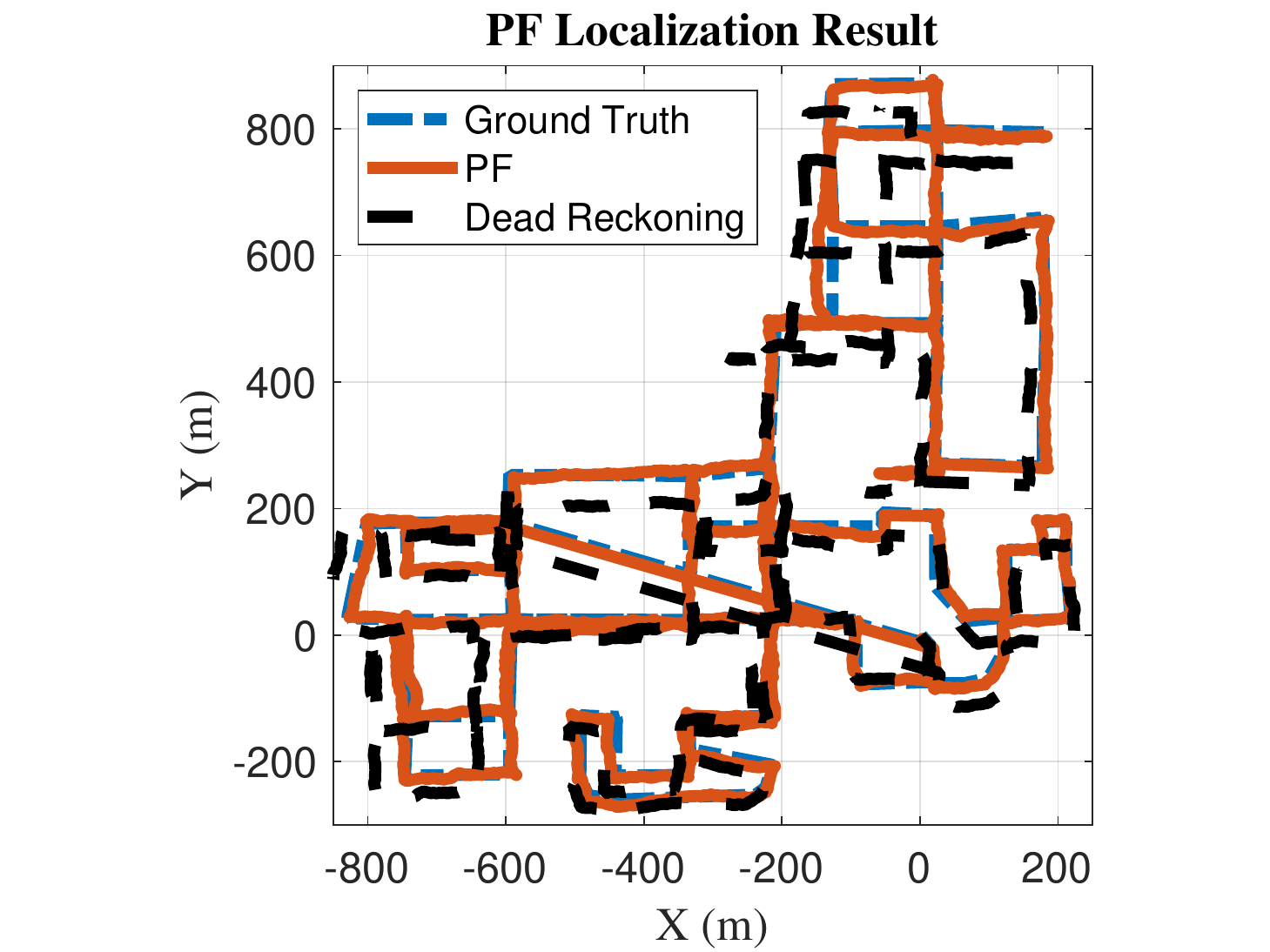}}\label{fig:CAPF.a}}
    \qquad
    \subfloat[Error in X coordinate.]{{\includegraphics[width=0.4\columnwidth]{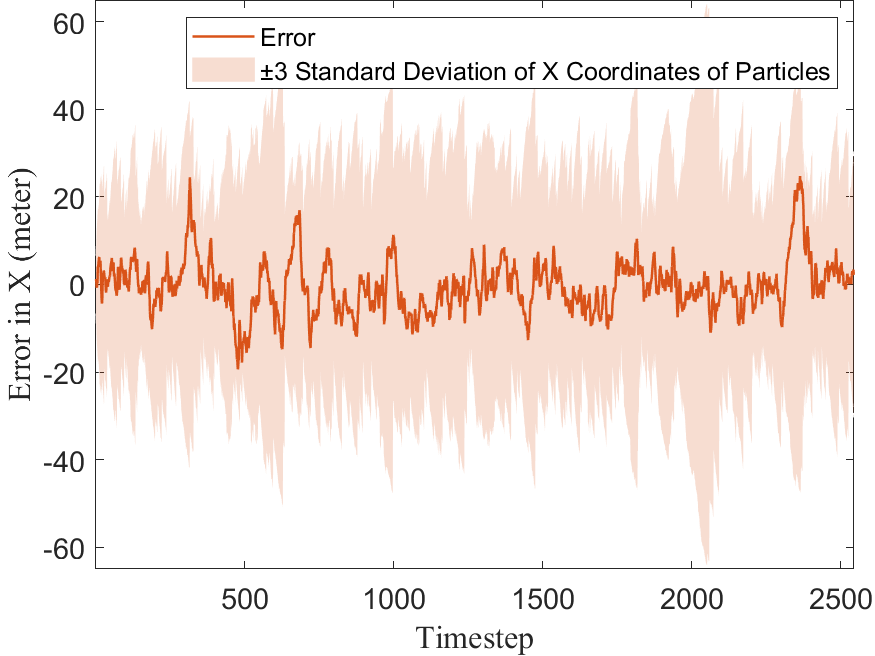} }\label{fig:CAPF.b}}
    \qquad
    \subfloat[Error in Y coordinate.]{{\includegraphics[width=0.4\columnwidth]{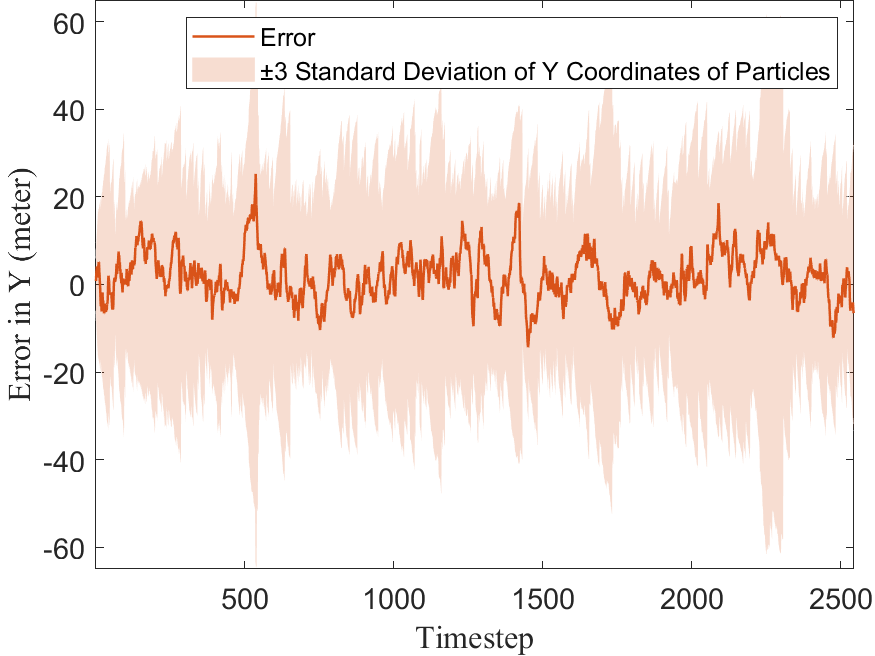}}}
    \caption{\DD{Localization performance for CAPF pipeline. Error in coordinate estimation at every time step is also shown.}}%
    \label{fig:CAPF}%
 \end{figure}

\begin{figure}
    \centering
     \subfloat[Plot of ground truth, dead reckoning and particle filter trajectories]{{\includegraphics[width=0.9\columnwidth]{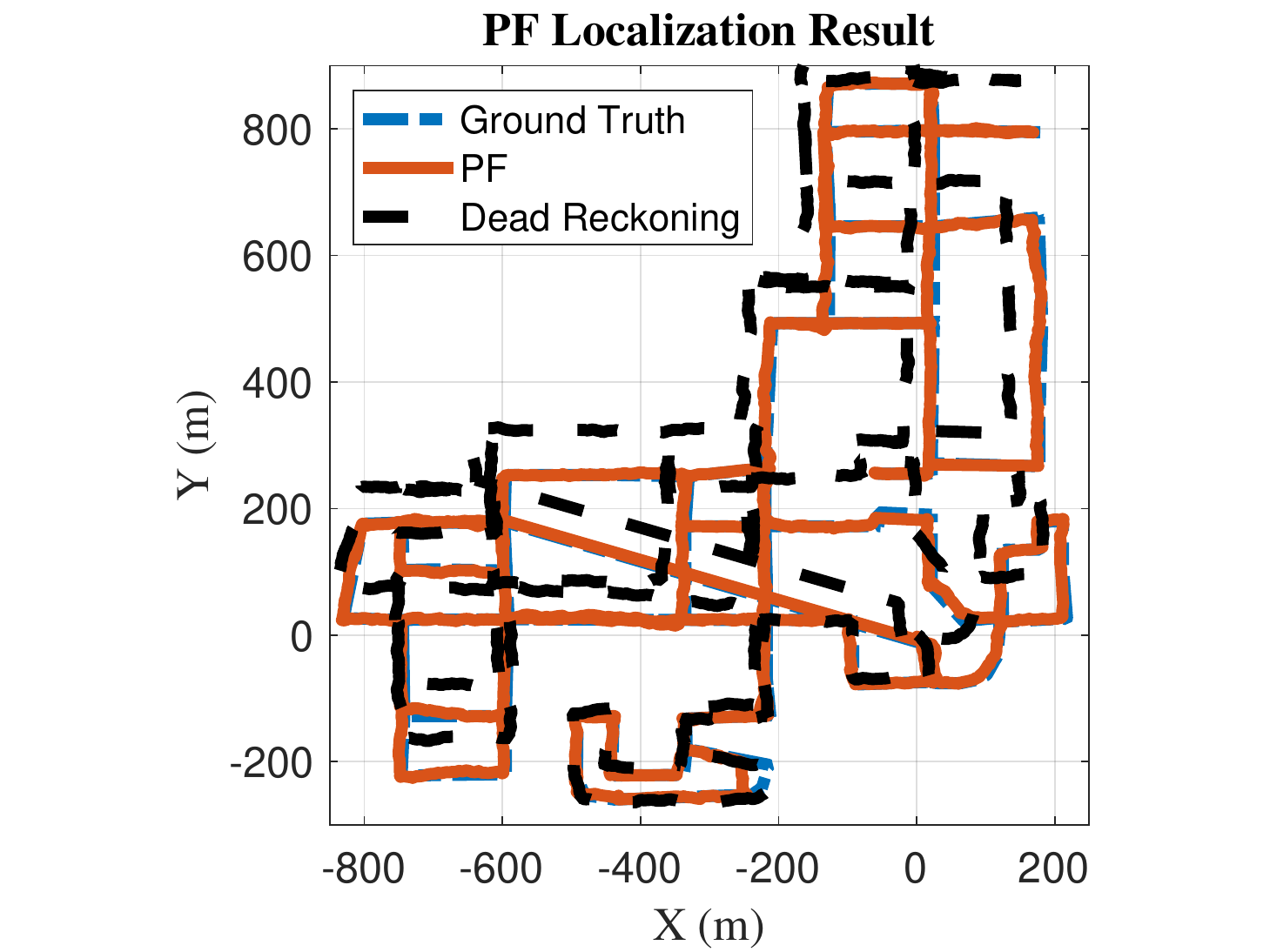} }\label{fig:PPF.a}}
    \qquad
    \subfloat[Error in X coordinate.]{{\includegraphics[width=0.4\columnwidth]{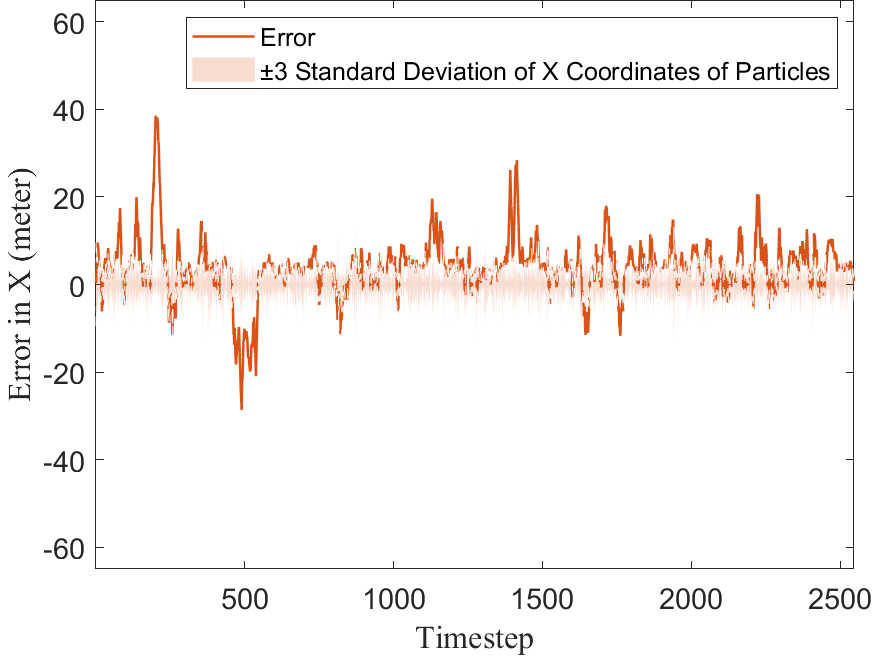} }\label{fig:PPF.b}}
    \qquad
    \subfloat[Error in Y coordinate.]{{\includegraphics[width=0.4\columnwidth]{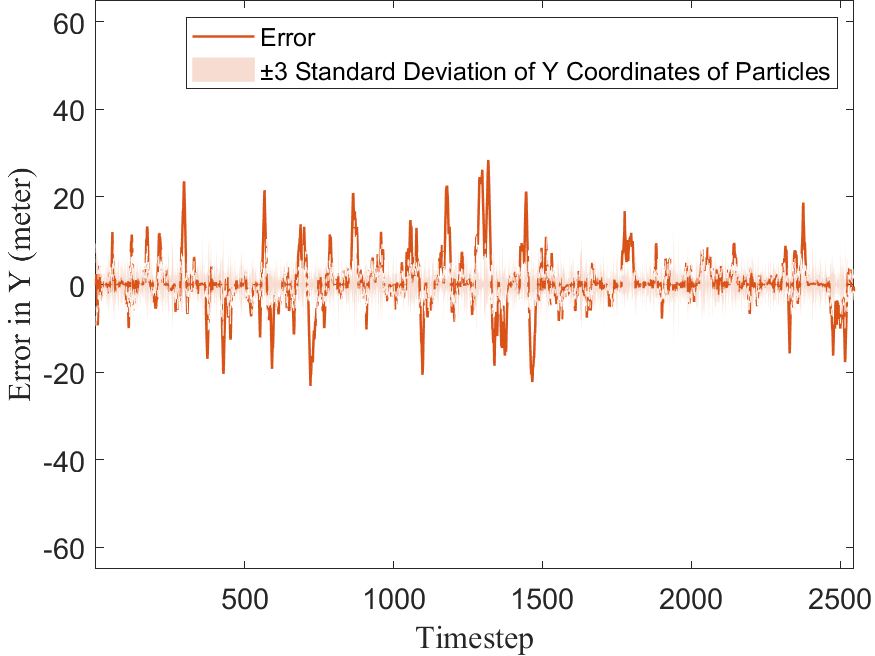}}}
    \caption{\DD{Localization performance for PPF pipeline. Error in coordinate estimation at every time step is also shown.}}%
    \label{fig:PPF}%
 \end{figure}

 We initialize the particle filter by sampling 200 particles around the initial location \DD{(known to the robot)} from a Gaussian distribution with a standard deviation of 4\textit{m}. The ground truth velocity obtained from the IMU data is artificially corrupted by a diagonal covariance matrix of standard deviation 1\textit{m} to simulate real-world noise. \DD{We use this velocity to estimate the dead reckoning trajectory.} This is used for propagating the particles in the update step. 
 As described in Section~\ref{sec:methodology}, PPF and CAPF methodologies primarily differ in the weighting scheme for the particles. 
The covariance matrix used to model the noise in the measurements is $\Sigma_{o}=diag([ 3,3 ]^{T})$. Figure~\ref{fig:CAPF} shows the performance of CAPF over the AirSim dataset while Figure~\ref{fig:PPF} shows the performance of PPF with top 20 predictions. It can be observed from Figures~\ref{fig:CAPF.a},~\ref{fig:CAPF.b},~\ref{fig:PPF.a} and \ref{fig:PPF.b} that PPF was very close to CAPF in terms of localization error as shown in Table~\ref{tab:ppf v/s capf} but in CAPF the error was more consistent with the $\pm$3 standard deviation bound.
{\paragraph*{Initial Estimate} We also experimented with relaxing the assumption that the initial location of the ground robot is known for the CAPF strategy. We start by predicting the first location using the same methodology used in PPF. We retrieve the top 20 neighbors from the aerial dataset for the first query ground view image. Then, we find the mean of the particles in the largest cluster after mean shift clustering and use it to initialize the particles. We increased the variance with which the particles are sampled to 9\textit{m} considering the noise in the initial estimate. We observed that the localization performance over the complete dataset largely remained unchanged giving a localization error of 6.8$\pm$4.3 when dead reckoning error was 37.6$\pm$17.5. On the test dataset, the localization error increased to 19.5$\pm$17.12\textit{m} with a dead reckoning error of 70.3$\pm$9.3\textit{m}.}

 \begin{table}
\centering
\caption{Evaluation of the PPF v/s CAPF methodology}
\label{tab:PPF v/s CAPF}
\begin{tabular}{|l|l|l|} 
\hline
\bf PF & \multicolumn{2}{c|}{\bf Localization Error $\pm$ Standard deviation (\textit{m}) }  \\ 
\hline
                   & Complete Dataset              & Test Dataset                      \\ 
\cline{2-3}
\bf PPF                & 8.24$\pm$5.90 & 10.70$\pm$8.86   \\
\bf CAPF               & \bf6.89$\pm$4.33 & \bf9.70$\pm$5.54    \\
\hline
\end{tabular}
\label{tab:ppf v/s capf}
\end{table}

\subsubsection{Altitude}
We performed five  experiments to analyze the effect of the altitude of aerial image collection on localization accuracy. For each altitude, we trained a separate network for cross-view matching and then used the aerial and ground view image descriptors generated by this network to do the particle filter localization. It is important to assess the trade-off between altitude and cross-view localization performance as it might not always be possible to fly the aerial robot at a given altitude. \DD{Localization error reported throughout this section refers to the L2 norm between the predicted and ground truth position.}

\begin{figure}
      \centering
      \subfloat[Recall accuracy on test data for different altitudes.]{{\includegraphics[width=0.4\columnwidth]{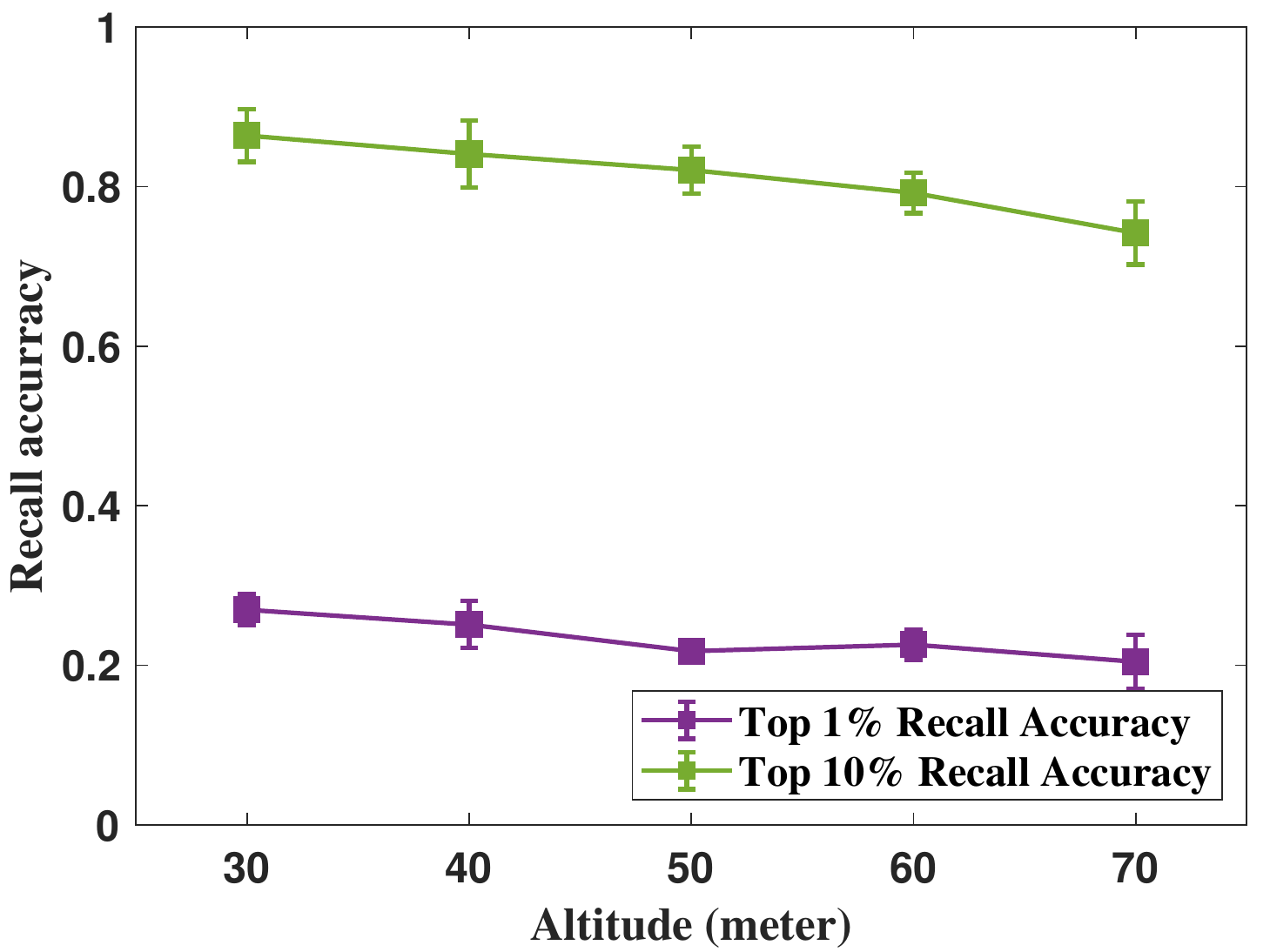} }\label{fig:test accuracy across height}}
    \qquad
    \subfloat[ \DD{Recall accuracy on test data for different FOVs.}]{{\includegraphics[width=0.4\columnwidth]{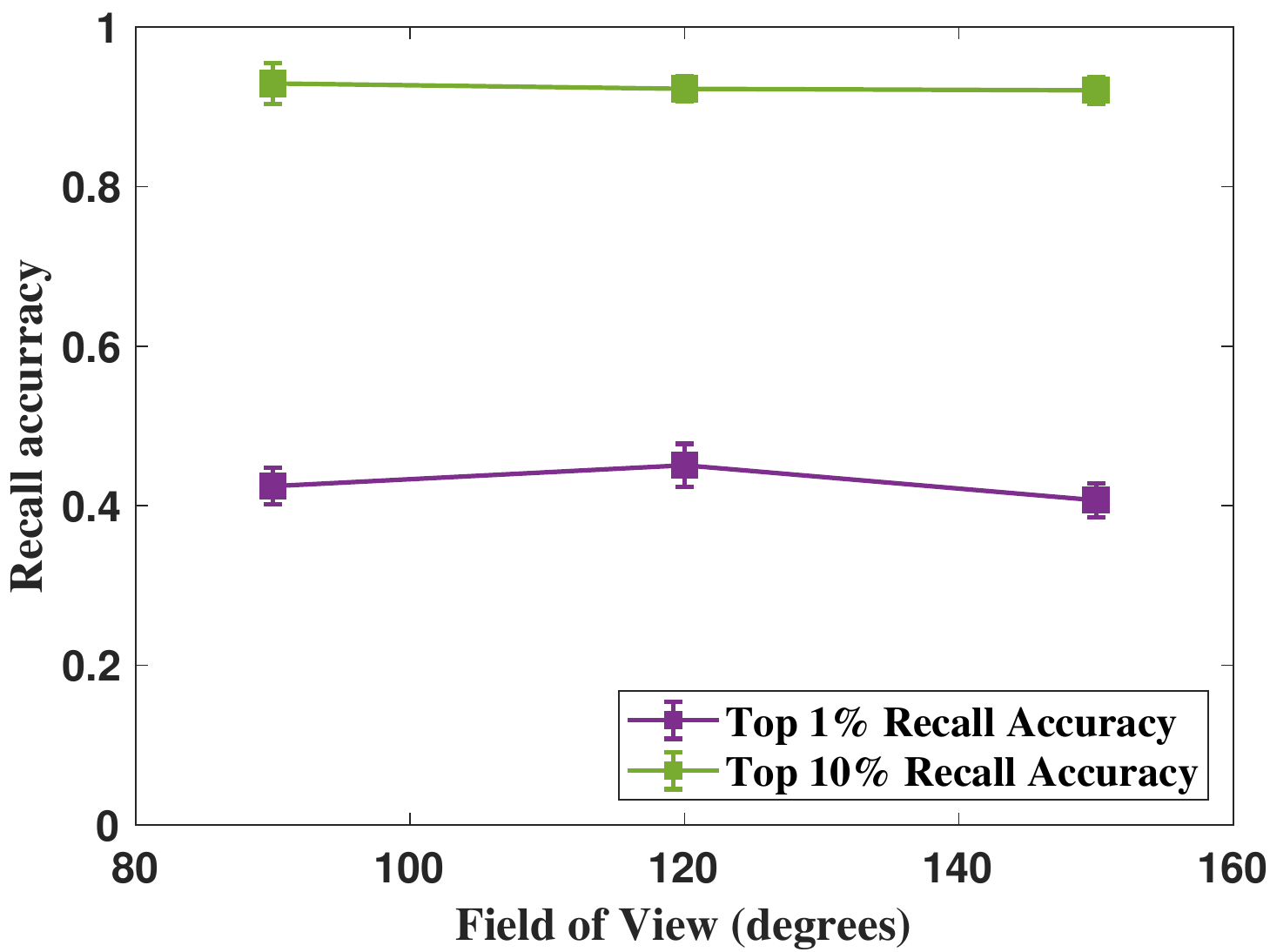} }\label{fig: test accuracy FOV}}
    \caption{Comparison of recall accuracy for test data across different altitudes and FOVs. The FOV of 150$\degree$ and -90$\degree$ pitch is kept consistent across all test settings for evaluation across altitudes. The altitude of 50\textit{m} and -50$\degree$ pitch is consistent across all the test settings for FOV evaluation.}%
    \label{fig:retrieval}%
    \label{fig:recall_accuracy}
\end{figure}

\begin{figure}
    \centering
    \subfloat[Complete dataset.]{{\includegraphics[width=0.4\columnwidth]{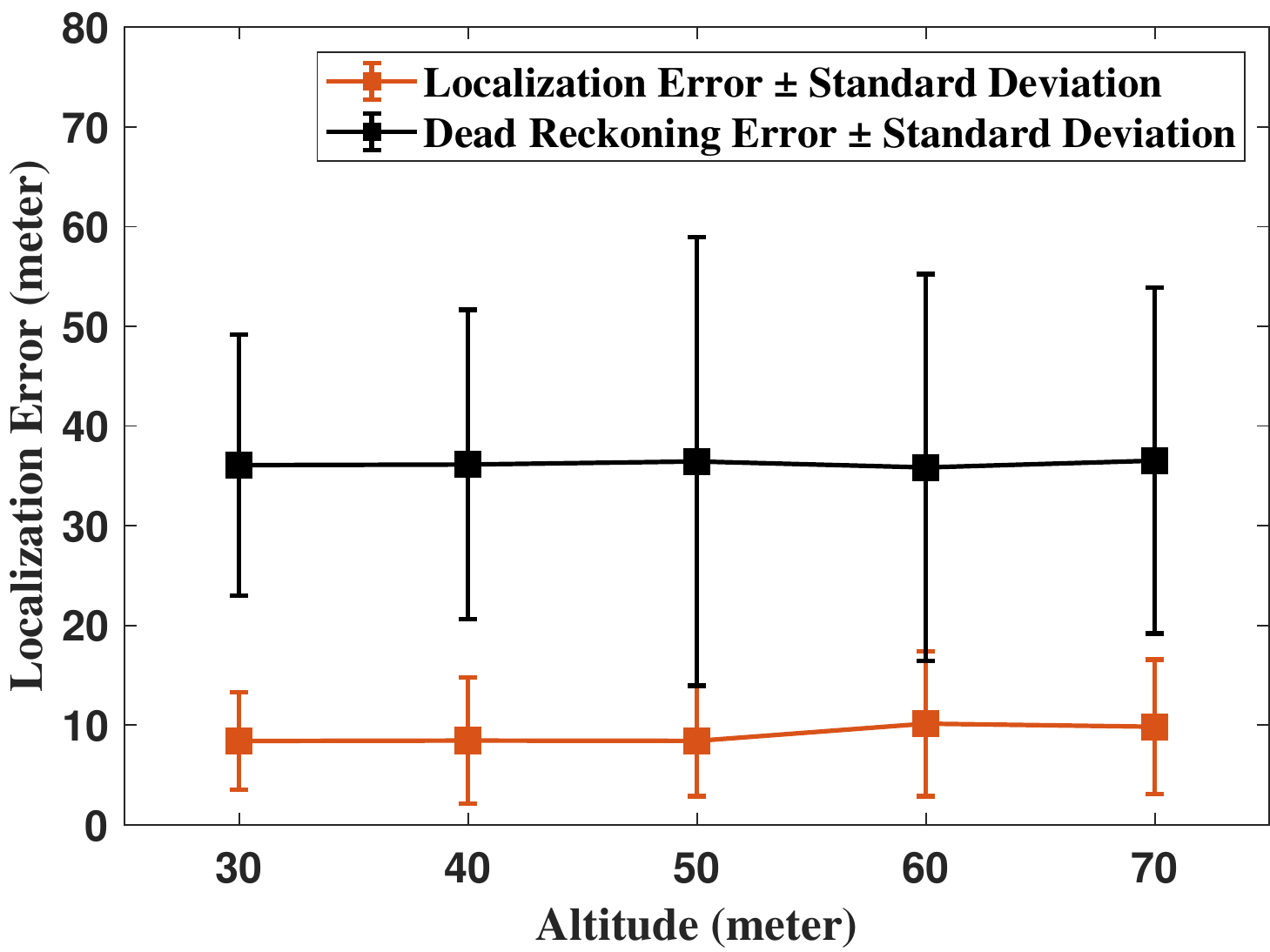} }\label{fig:height train}}
    \qquad
    \subfloat[Test dataset.]{{\includegraphics[width=0.4\columnwidth]{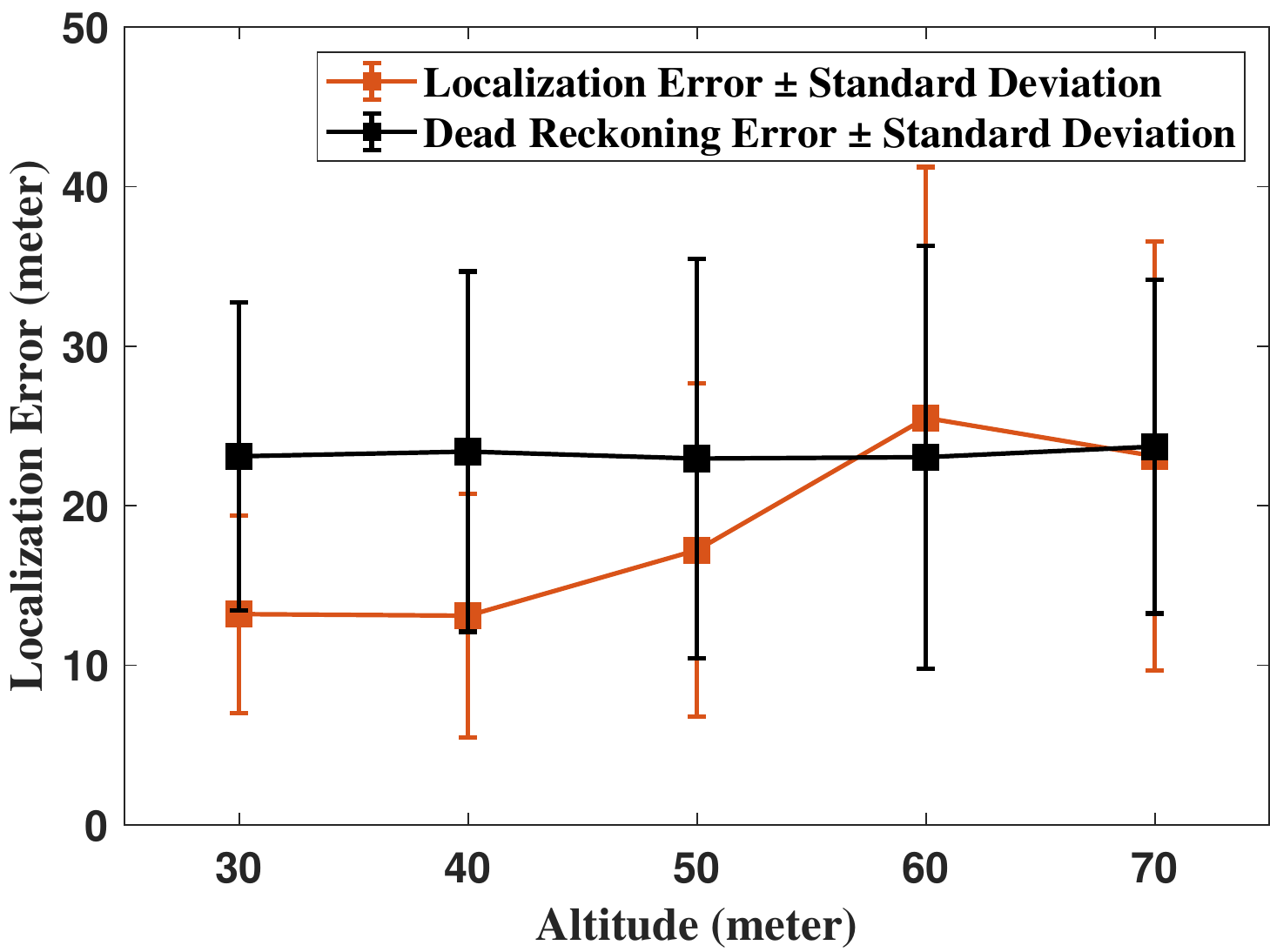} }\label{fig:height test}}
    \caption{Plot for localization error (L2 norm) and dead reckoning for comparison of particle filter performance across different altitudes. All the values are averaged over 5 runs of particle filter.}%
    \label{fig:Error v/s altitude}%
 \end{figure}

It can be seen from Figure~\ref{fig:test accuracy across height} that the top 1\% recall accuracy was highest for an altitude of 30\textit{m}. This is the accuracy of individual measurements. However, the localization error is approximately the same for both complete and test datasets as shown in Figure~\ref{fig:height train} and Figure~\ref{fig:height test}, respectively. The results presented are averages over five runs of the particle filter.


\subsubsection{Field of View}
\begin{figure}
    \centering
    \subfloat[Complete dataset. ]{{\includegraphics[width=0.4\columnwidth]{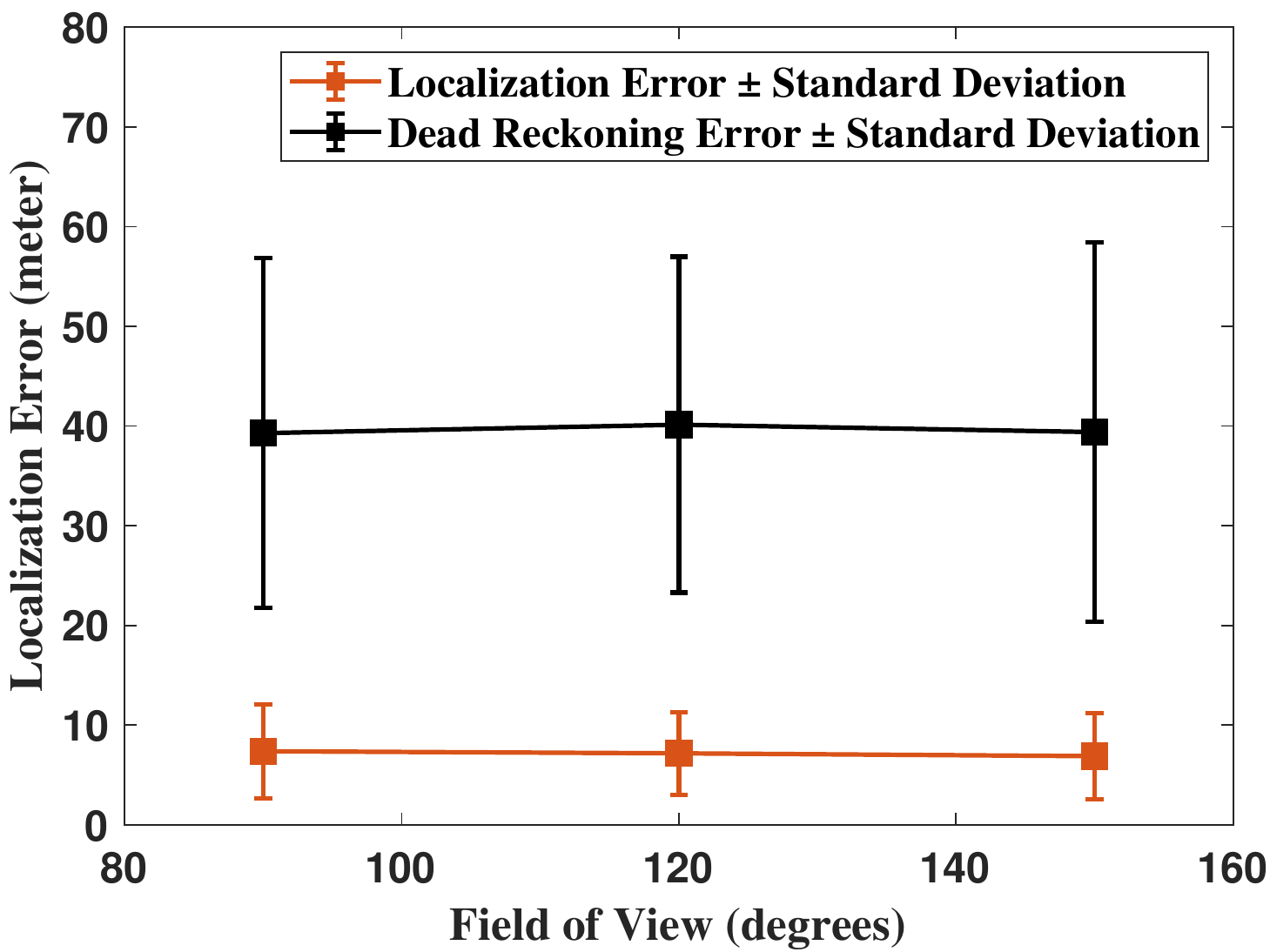} }\label{fig:lfov.a}}
    \qquad
    \subfloat[Test dataset.  ]{{\includegraphics[width=0.4\columnwidth]{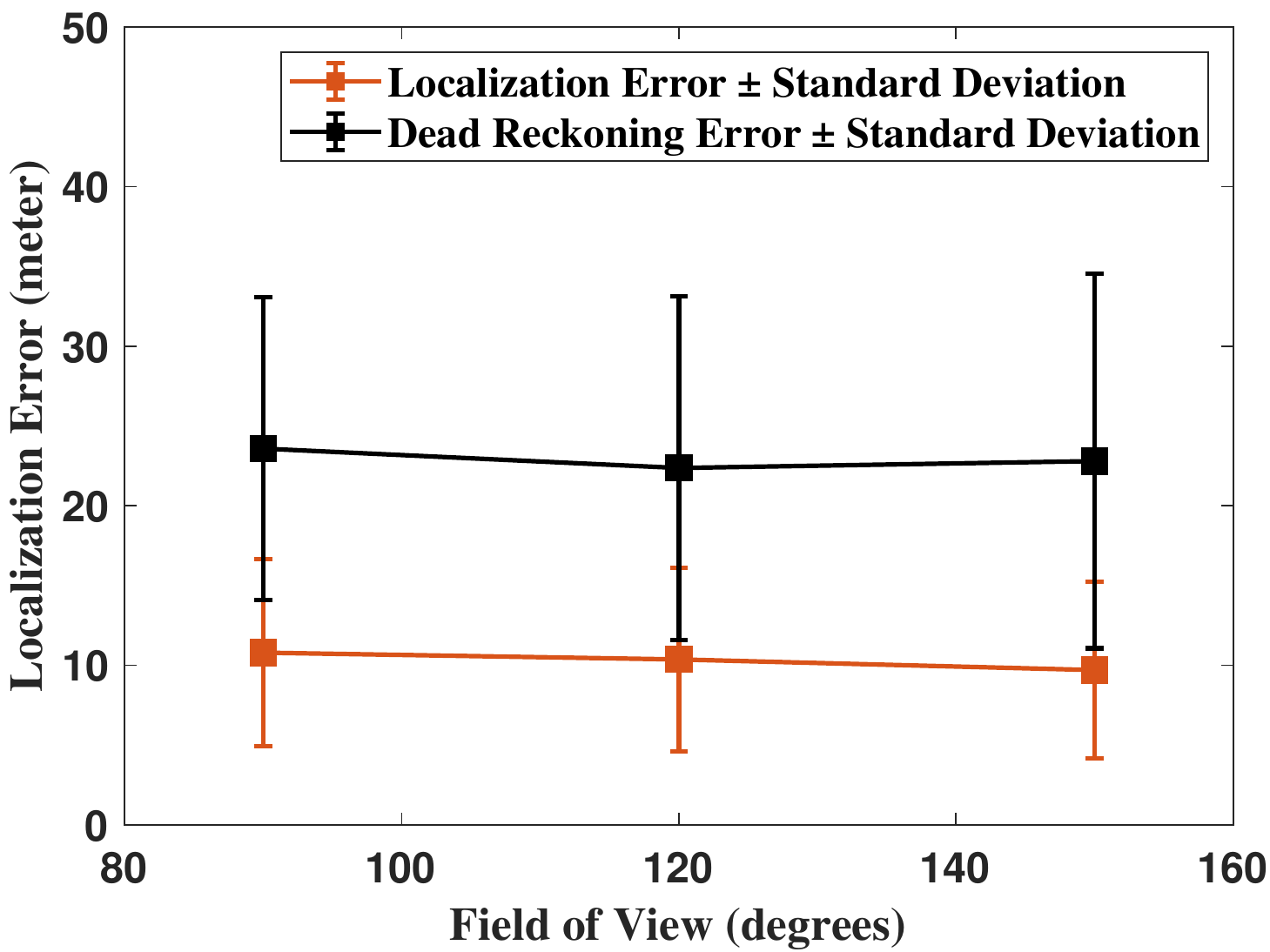}}\label{fig:lfov.b}}
    \caption{Plot for localization error (L2 norm) and dead reckoning for comparison of particle filter performance across different FOV. Values shown are averaged over 5 runs of particle filter.}%
    \label{fig:Error v/s FOv}%
 \end{figure}

FOV of the ground vehicle also plays an important role in cross-view scene understanding. Thus, we analyzed the performance of our localization pipeline across three different FOVs and assessed the trade-off between them in terms of recall accuracy and localization error. The performance of all of these FOVs is analyzed for an altitude of 50\textit{m}. For 1\% recall accuracy, the 120$\degree$ FOV performed the best as shown in Figure~\ref{fig: test accuracy FOV}. \DD{Regardless}, when it came to the localization errors, all three FOV performed approximately the same, as seen in Figure~\ref{fig:lfov.a} and Figure~\ref{fig:lfov.b}.

 \subsubsection{Pitch}
 \begin{figure}[!htpb]
      \centering
      \includegraphics[width=0.6\columnwidth]{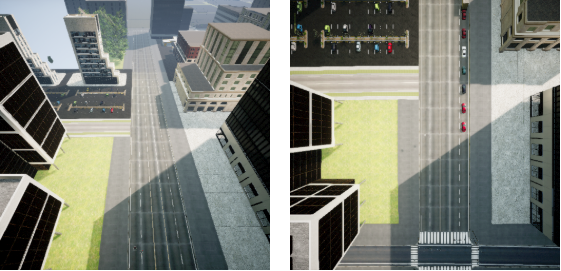}
      \caption{Images collected with pitch of -50 (left image) and pitch of -90 (right image) at an altitude of 50 meter.}
      \label{fig:pitch}
\end{figure}
 Changing the pitch of the camera mount is a small adjustment that can change the amount of information contained in an image (\DD{as shown in Figure~\ref{fig:pitch})} and thus the performance of both retrieval and localization. This hypothesis was justified by our experiments conducted for two different pitch values for an altitude of 50\textit{m}. Changing the pitch from -90$\degree$ (top-down) to -50$\degree$ (look-ahead) has a significant impact in the retrieval performance as seen from the Table~\ref{tab:pitch}. This also resulted in better localization performance.

\begin{table}
\setlength\tabcolsep{2pt}
\centering
\caption{Evaluation of pitch for the aerial camera}
\begin{tabular}{|l|ll|ll|} 
\hline
\bf Pitch                & \multicolumn{2}{c|}{\bf Test Accuracy}&  \multicolumn{2}{c|}{\bf Localization Error $\pm$ Standard Deviation(\textit{m})}                                   \\ 
\hline
                     & \multicolumn{1}{l|}{\begin{tabular}[c]{@{}l@{}}Top 1\% \\Recall\end{tabular}} & \begin{tabular}[c]{@{}l@{}}Top 10\% \\Recall\end{tabular} & \multicolumn{1}{l|}{\begin{tabular}[c]{@{}l@{}}~~~~~~~Complete~~~~~~~ \\~~~~~~~~Dataset~~~~~~~\end{tabular}} & \begin{tabular}[c]{@{}l@{}}~~~~~~~Test~~~~~ \\~~~~~~Dataset~~~~\end{tabular}  \\ 
\cline{2-5}
\bf-50                  & \multicolumn{1}{l|}{0.3229}                                                   & 0.9208                                                   & \multicolumn{1}{l|}{~~~~~~\bf6.89$\pm$4.33}                                                 & ~~~~\bf9.70$\pm$5.54                                              \\
\bf-90                  & \multicolumn{1}{l|}{0.1775}                                                   & 0.7499                                                    & \multicolumn{1}{l|}{~~~~~~8.42$\pm$5.55~}                                                & ~~17.22$\pm$10.44                                            \\ 
\hline
\end{tabular}
\label{tab:pitch}
\end{table}
\subsubsection{Evaluation on Real-World Data}
\begin{figure}
    \centering
    \subfloat[Interleaved samples. ]{{\includegraphics[width=0.4\columnwidth]{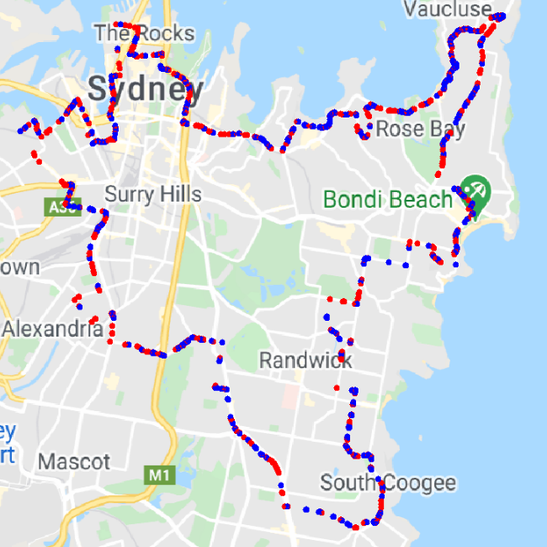} }\label{fig:train_test_samples}}
    \qquad
    \subfloat[Separate samples.  ]{{\includegraphics[width=0.4\columnwidth]{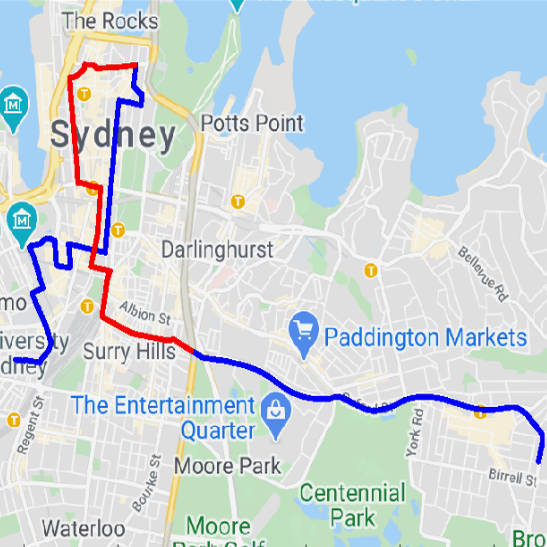}}\label{fig:dense}}
    \caption{Training (blue) and Testing (red) samples. On the left, samples are randomly chosen from the entire trajectory while on the right, train and test samples are separated into two different trajectories.}%
    \label{fig:Real world samples}%
 \end{figure}

\SV{We also generated a real-world dataset to evaluate the performance of our framework. Specifically, we collected aerial and Street View image pairs from Sydney, Australia, by charting a specific trajectory in Google My Maps~\cite{gmaps}. The KML (Keyhole Markup Language) file generated as a result, contains ordered list of GPS coordinates which was then used to query aerial images with a pitch of $45^{o}$ \cite{gorelick2017google}, as well as corresponding Street View images. We concatenated 4 Street View images with different orientations together to get the query ground view panorama.

We conduct two experiments with this dataset. In the first case, out of the $1569$ image pairs that we collect along a trajectory of length $45.54$\textit{km}, $1099$ of them are randomly sampled to form the training dataset while the remaining \DD{$470$} pairs are used as the test dataset (Figure~\ref{fig:train_test_samples}). The entire trajectory encompasses not only feature rich samples from narrow and congested lanes but also from sparse regions with less number of features.}

\SV{We used CVM-Net, trained on CVUSA dataset \cite{workman2015localize}, and fine-tuned this network by training only the final NetVLAD layer with the Google dataset as input. We train the model for $120$ epochs with a batch size of $12$ image pairs and a learning rate of $10^{-4}$. The test accuracy we obtain at the end of 120 epochs \DD{is} $\mathbf{92.8\%}$ for top $10\%$ recall accuracy and $\mathbf{54.8\%}$ for top $1\%$ recall accuracy. }

\begin{figure}
      \centering
      \includegraphics[width=0.7\columnwidth]{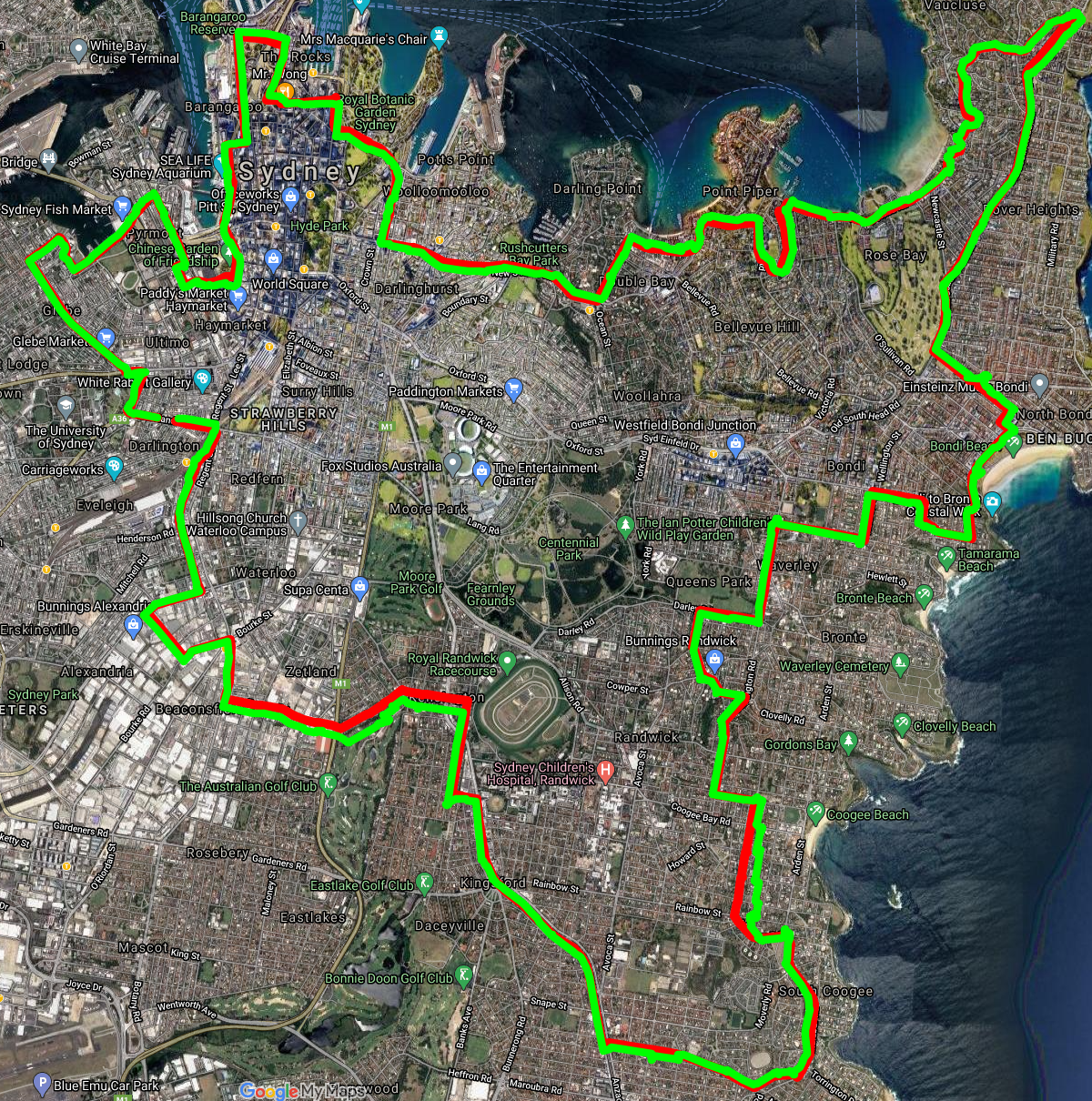}
      \caption{Final trajectory estimated using CAPF approach on google dataset collected in the city of Sydney, Australia. Estimated trajectory is shown in green while the ground truth is marked in red. }
      \label{fig:google maps}
\end{figure}

\DD{We use CAPF to evaluate the localization performance on this dataset. The velocity measurements are simulated on the trajectory by using the GPS coordinates (from the geotagged images) after which the GPS data is corrupted with diagonal noise with standard deviation of 3\textit{m}. We increased the number of particles in the particle filter to 1000, since the number of images collected are sparse with respect to the length of the trajectory. We observed a localization error of 46.5$\pm$34.3\textit{m} with a dead reckoning error of 155.8$\pm$74.8\textit{m} averaged over 8 runs of the particle filter. A sample result is shown in Figure~\ref{fig:google maps}. We also experimented with simulating even noisier velocity measurements by increasing the standard deviation of noise to 5\textit{m}. This increased the dead reckoning error to 442.0$\pm$193.5\textit{m} but the localization error was only increased to 72.8$\pm$45.9\textit{m}.}


\DD{We also test how the CAPF methodology generalizes to unseen areas. We completely separated the trajectories used for training and testing, but they are still from the same area (Figure. \ref{fig:dense}). The entire trajectory is 12.7\textit{km} long and the test trajectory is 4.14\textit{km} long. Using the same parameters mentioned above and a noise with standard deviation of 3\textit{m} in the simulated velocity measurements, we observed an error of 153.8$\pm$104.3\textit{m} on complete trajectory (test+train) with dead reckoning error 130.2$\pm$78.6. We observed an error of 117.7$\pm$81.7\textit{m} on just the test trajectory with dead reckoning error 172.3$\pm$136.9. This suggests potential future work that generalizes better to unseen environments.}


\section{DISCUSSION}
\label{sec:conclusion}

We investigate the performance of cross-view matching for localization of a moving vehicle using an aerial image database. Prior work had shown that cross-view matching techniques, such as CVM-Net, can successfully retrieve aerial images that are closest to a given query ground view image. We show how this can be used, along with a particle filter, to improve the localization of a ground vehicle. In our experiments, cross-view matching was the only perception module used. Nonetheless, in practice, one would combine cross-view matching along with onboard perception.

We evaluated CVM-Net through \DD{four} design choices. We have the following conclusions: (1) We find that instead of choosing the top $k$ nearest neighbors (PPF), using all the images in the aerial database to weigh the particles (CAPF) performs better. The localization error is similar but the consistency of the latter is better. We conjecture that this is due to the susceptibility of the top $k$ retrievals to outliers. (2) We find that although the retrieval accuracy improves as the altitude of aerial images decreases, the localization performance over a trajectory is unaffected. We believe this is because the retrieval accuracy only depends on the top $k$ neighbors, whereas the localization accuracy depends on how close the global descriptors are. Along a trajectory, we expect similar aerial views that, in turn, have similar global descriptors. Therefore, in the CAPF approach, they will give similar weights to nearby particles. On the other hand, unless you select the exact image from the top, the retrieval accuracy will be hampered. We believe this is why although the retrieval accuracy decreases with increasing altitude, the localization performance is largely unaffected. (3) Similar conclusions can be reached for FOV. Higher FOV leads to better retrieval (only marginally) but similar localization performance. (4) On the other hand, the pitch of the aerial images has a significant impact. Top-down aerial images perform poorly as compared to front-facing ones. 

\DD{We note that for establishing the ground truth for the retrieval performance we also need to consider the overlap between images. Currently, the ground truth definition for our recall accuracy plots~\ref{fig:recall_accuracy} comes by pairing ground/aerial images during the data collection process itself. However, a single ground view image can correspond to multiple aerial views as the height increases. Further work is required to evaluate this scenario. Nevertheless, we conjecture that overlapping aerial views will not affect the performance of the localization pipeline since in CAPF the performance does not rely on a single ground truth aerial view image. Even with overlapping views, the descriptors learned by CVM-Net for those views will be similar. Consequently, the CAPF method will assign similar weights to those descriptors. Also, in PPF the preprocessing step of mean shift clustering will cluster the overlapping views together with the ground truth due to their similarity keeping the noisy measurements approximately the same. Thus, retrieval performance might vary depending on the definition of ground due to overlap we expect the localization performance to stay the same.}

We expect that the observations from this paper can eliminate some of the guesswork in deploying cross-view matching for localization. An immediate avenue for future work is to evaluate this through field experiments. Since cross-view matching has been extensively evaluated with real-world data, we expect the results to be similar. 

\bibliographystyle{IEEEtran}
\bibliography{IEEEabrv,ref.bib}

\begin{thebibliography}{10}
\providecommand{\url}[1]{#1}
\csname url@samestyle\endcsname
\providecommand{\newblock}{\relax}
\providecommand{\bibinfo}[2]{#2}
\providecommand{\BIBentrySTDinterwordspacing}{\spaceskip=0pt\relax}
\providecommand{\BIBentryALTinterwordstretchfactor}{4}
\providecommand{\BIBentryALTinterwordspacing}{\spaceskip=\fontdimen2\font plus
\BIBentryALTinterwordstretchfactor\fontdimen3\font minus
  \fontdimen4\font\relax}
\providecommand{\BIBforeignlanguage}[2]{{%
\expandafter\ifx\csname l@#1\endcsname\relax
\typeout{** WARNING: IEEEtran.bst: No hyphenation pattern has been}%
\typeout{** loaded for the language `#1'. Using the pattern for}%
\typeout{** the default language instead.}%
\else
\language=\csname l@#1\endcsname
\fi
#2}}
\providecommand{\BIBdecl}{\relax}
\BIBdecl

\bibitem{hu2018cvm}
S.~Hu, M.~Feng, R.~M. Nguyen, and G.~Hee~Lee, ``Cvm-net: Cross-view matching
  network for image-based ground-to-aerial geo-localization,'' in
  \emph{Proceedings of the IEEE Conference on Computer Vision and Pattern
  Recognition}, 2018, pp. 7258--7267.

\bibitem{whyatt2008noisy}
J.~D. Whyatt, G.~Davies, M.~Walker, C.~G. Pooley, P.~Coulton, and W.~Bamford,
  ``Noisy school kids: using gps in an urban environment.'' \emph{GISRUK 2008},
  2008.

\bibitem{Castaldo_2015_ICCV_Workshops}
F.~Castaldo, A.~Zamir, R.~Angst, F.~Palmieri, and S.~Savarese, ``Semantic
  cross-view matching,'' in \emph{The IEEE International Conference on Computer
  Vision (ICCV) Workshops}, December 2015.

\bibitem{zhu2020leveraging}
Q.~Zhu, Z.~Wang, H.~Hu, L.~Xie, X.~Ge, and Y.~Zhang, ``Leveraging
  photogrammetric mesh models for aerial-ground feature point matching toward
  integrated 3d reconstruction,'' \emph{arXiv preprint arXiv:2002.09085}, 2020.

\bibitem{workman2015wide}
S.~Workman, R.~Souvenir, and N.~Jacobs, ``Wide-area image geolocalization with
  aerial reference imagery,'' in \emph{Proceedings of the IEEE International
  Conference on Computer Vision}, 2015, pp. 3961--3969.

\bibitem{viswanathan2014vision}
A.~Viswanathan, B.~R. Pires, and D.~Huber, ``Vision based robot localization by
  ground to satellite matching in gps-denied situations,'' in \emph{2014
  IEEE/RSJ International Conference on Intelligent Robots and Systems}.\hskip
  1em plus 0.5em minus 0.4em\relax IEEE, 2014, pp. 192--198.

\bibitem{shan2014accurate}
Q.~Shan, C.~Wu, B.~Curless, Y.~Furukawa, C.~Hernandez, and S.~M. Seitz,
  ``Accurate geo-registration by ground-to-aerial image matching,'' in
  \emph{2014 2nd International Conference on 3D Vision}, vol.~1.\hskip 1em plus
  0.5em minus 0.4em\relax IEEE, 2014, pp. 525--532.

\bibitem{lin2013cross}
T.-Y. Lin, S.~Belongie, and J.~Hays, ``Cross-view image geolocalization,'' in
  \emph{Proceedings of the IEEE Conference on Computer Vision and Pattern
  Recognition}, 2013, pp. 891--898.

\bibitem{lin2015learning}
T.-Y. Lin, Y.~Cui, S.~Belongie, and J.~Hays, ``Learning deep representations
  for ground-to-aerial geolocalization,'' in \emph{Proceedings of the IEEE
  conference on computer vision and pattern recognition}, 2015, pp. 5007--5015.

\bibitem{liu2019lending}
L.~Liu and H.~Li, ``Lending orientation to neural networks for cross-view
  geo-localization,'' \emph{arXiv preprint arXiv:1903.12351}, 2019.

\bibitem{koch2015siamese}
G.~Koch, R.~Zemel, and R.~Salakhutdinov, ``Siamese neural networks for one-shot
  image recognition,'' in \emph{ICML deep learning workshop}, vol.~2.\hskip 1em
  plus 0.5em minus 0.4em\relax Lille, 2015.

\bibitem{Fox2001ParticleFF}
D.~Fox, S.~Thrun, W.~Burgard, and F.~Dellaert, ``Particle filters for mobile
  robot localization,'' in \emph{Sequential Monte Carlo Methods in Practice},
  2001.

\bibitem{DBLP:journals/corr/abs-1809-05477}
\BIBentryALTinterwordspacing
L.~Heng, B.~Choi, Z.~Cui, M.~Geppert, S.~Hu, B.~Kuan, P.~Liu, R.~M.~H. Nguyen,
  Y.~C. Yeo, A.~Geiger, G.~H. Lee, M.~Pollefeys, and T.~Sattler, ``Project
  autovision: Localization and 3d scene perception for an autonomous vehicle
  with a multi-camera system,'' \emph{CoRR}, vol. abs/1809.05477, 2018.
  [Online]. Available: \url{http://arxiv.org/abs/1809.05477}
\BIBentrySTDinterwordspacing

\bibitem{hu2020image}
S.~Hu and G.~H. Lee, ``Image-based geo-localization using satellite imagery,''
  \emph{International Journal of Computer Vision}, vol. 128, no.~5, pp.
  1205--1219, 2020.

\bibitem{Doan2019VisualLU}
A.-D. Doan, Y.~Latif, T.-J. Chin, Y.~Liu, S.~F. Ch'ng, T.-T. Do, and I.~Reid,
  ``Visual localization under appearance change: A filtering approach,''
  \emph{2019 Digital Image Computing: Techniques and Applications (DICTA)}, pp.
  1--8, 2019.

\bibitem{DBLP:journals/corr/ShahDLK17}
\BIBentryALTinterwordspacing
S.~Shah, D.~Dey, C.~Lovett, and A.~Kapoor, ``Airsim: High-fidelity visual and
  physical simulation for autonomous vehicles,'' \emph{CoRR}, vol.
  abs/1705.05065, 2017. [Online]. Available:
  \url{http://arxiv.org/abs/1705.05065}
\BIBentrySTDinterwordspacing

\bibitem{workman2015localize}
S.~Workman, R.~Souvenir, and N.~Jacobs, ``Wide-area image geolocalization with
  aerial reference imagery,'' in \emph{IEEE International Conference on
  Computer Vision (ICCV)}, 2015, pp. 1--9, acceptance rate: 30.3\%.

\bibitem{zheng2020university}
Z.~Zheng, Y.~Wei, and Y.~Yang, ``University-1652: A multi-view multi-source
  benchmark for drone-based geo-localization,'' \emph{ACM Multimedia}, 2020.

\bibitem{tian2017cross}
Y.~Tian, C.~Chen, and M.~Shah, ``Cross-view image matching for geo-localization
  in urban environments,'' in \emph{Proceedings of the IEEE Conference on
  Computer Vision and Pattern Recognition}, 2017, pp. 3608--3616.

\bibitem{hays2008im2gps}
J.~Hays and A.~A. Efros, ``Im2gps: estimating geographic information from a
  single image,'' in \emph{2008 ieee conference on computer vision and pattern
  recognition}.\hskip 1em plus 0.5em minus 0.4em\relax IEEE, 2008, pp. 1--8.

\bibitem{agarwal2009building}
S.~Agarwal, N.~Snavely, I.~Simon, S.~M. Seitz, and R.~Szeliski, ``Building rome
  in a day,'' in \emph{2009 IEEE 12th international conference on computer
  vision}.\hskip 1em plus 0.5em minus 0.4em\relax IEEE, 2009, pp. 72--79.

\bibitem{irschara2009structure}
A.~Irschara, C.~Zach, J.-M. Frahm, and H.~Bischof, ``From structure-from-motion
  point clouds to fast location recognition,'' in \emph{2009 IEEE Conference on
  Computer Vision and Pattern Recognition}.\hskip 1em plus 0.5em minus
  0.4em\relax IEEE, 2009, pp. 2599--2606.

\bibitem{li2010location}
Y.~Li, N.~Snavely, and D.~P. Huttenlocher, ``Location recognition using
  prioritized feature matching,'' in \emph{European conference on computer
  vision}.\hskip 1em plus 0.5em minus 0.4em\relax Springer, 2010, pp. 791--804.

\bibitem{sarlin2019coarse}
P.-E. Sarlin, C.~Cadena, R.~Siegwart, and M.~Dymczyk, ``From coarse to fine:
  Robust hierarchical localization at large scale,'' in \emph{Proceedings of
  the IEEE Conference on Computer Vision and Pattern Recognition}, 2019, pp.
  12\,716--12\,725.

\bibitem{kendall2015posenet}
A.~Kendall, M.~Grimes, and R.~Cipolla, ``Posenet: A convolutional network for
  real-time 6-dof camera relocalization,'' in \emph{Proceedings of the IEEE
  international conference on computer vision}, 2015, pp. 2938--2946.

\bibitem{imagenet_cvpr09}
J.~Deng, W.~Dong, R.~Socher, L.-J. Li, K.~Li, and L.~Fei-Fei, ``{ImageNet: A
  Large-Scale Hierarchical Image Database},'' in \emph{CVPR09}, 2009.

\bibitem{mapnet}
S.~Brahmbhatt, J.~Gu, K.~Kim, J.~Hays, and J.~Kautz, ``Geometry-aware learning
  of maps for camera localization,'' in \emph{IEEE Conference on Computer
  Vision and Pattern Recognition (CVPR)}, 2018.

\bibitem{doi:10.1177/0278364919839761}
\BIBentryALTinterwordspacing
S.~Garg, N.~Suenderhauf, and M.~Milford, ``Semantic–geometric visual place
  recognition: a new perspective for reconciling opposing views,'' \emph{The
  International Journal of Robotics Research}, vol.~0, no.~0, p.
  0278364919839761, 0. [Online]. Available:
  \url{https://doi.org/10.1177/0278364919839761}
\BIBentrySTDinterwordspacing

\bibitem{gawel2018x}
A.~Gawel, C.~Del~Don, R.~Siegwart, J.~Nieto, and C.~Cadena, ``X-view:
  Graph-based semantic multi-view localization,'' \emph{IEEE Robotics and
  Automation Letters}, vol.~3, no.~3, pp. 1687--1694, 2018.

\bibitem{synthia}
G.~Ros, L.~Sellart, J.~Materzynska, D.~Vazquez, and A.~Lopez, ``{The SYNTHIA
  Dataset}: A large collection of synthetic images for semantic segmentation of
  urban scenes,'' 2016.

\bibitem{wolf2005robust}
J.~Wolf, W.~Burgard, and H.~Burkhardt, ``Robust vision-based localization by
  combining an image-retrieval system with monte carlo localization,''
  \emph{IEEE transactions on robotics}, vol.~21, no.~2, pp. 208--216, 2005.

\bibitem{menegatti2004image}
E.~Menegatti, M.~Zoccarato, E.~Pagello, and H.~Ishiguro, ``Image-based monte
  carlo localisation with omnidirectional images,'' \emph{Robotics and
  Autonomous Systems}, vol.~48, no.~1, pp. 17--30, 2004.

\bibitem{xu2019robust}
S.~Xu, W.~Chou, and H.~Dong, ``A robust indoor localization system integrating
  visual localization aided by cnn-based image retrieval with monte carlo
  localization,'' \emph{Sensors}, vol.~19, no.~2, p. 249, 2019.

\bibitem{4155707}
W.~{Zhang} and J.~{Kosecka}, ``Image based localization in urban
  environments,'' in \emph{Third International Symposium on 3D Data Processing,
  Visualization, and Transmission (3DPVT'06)}, June 2006, pp. 33--40.

\bibitem{DBLP:journals/corr/RenHG015}
\BIBentryALTinterwordspacing
S.~Ren, K.~He, R.~B. Girshick, and J.~Sun, ``Faster {R-CNN:} towards real-time
  object detection with region proposal networks,'' \emph{CoRR}, vol.
  abs/1506.01497, 2015. [Online]. Available:
  \url{http://arxiv.org/abs/1506.01497}
\BIBentrySTDinterwordspacing

\bibitem{arandjelovic2013all}
R.~Arandjelovic and A.~Zisserman, ``All about vlad,'' in \emph{Proceedings of
  the IEEE conference on Computer Vision and Pattern Recognition}, 2013, pp.
  1578--1585.

\bibitem{Arandjelovi2016NetVLADCA}
R.~Arandjelovi{\'c}, P.~Gron{\'a}t, A.~Torii, T.~Pajdla, and J.~Sivic,
  ``Netvlad: Cnn architecture for weakly supervised place recognition,''
  \emph{2016 IEEE Conference on Computer Vision and Pattern Recognition
  (CVPR)}, pp. 5297--5307, 2016.

\bibitem{regmi2019bridging}
K.~Regmi and M.~Shah, ``Bridging the domain gap for ground-to-aerial image
  matching,'' in \emph{Proceedings of the IEEE International Conference on
  Computer Vision}, 2019, pp. 470--479.

\bibitem{shi2019spatial}
Y.~Shi, L.~Liu, X.~Yu, and H.~Li, ``Spatial-aware feature aggregation for image
  based cross-view geo-localization,'' in \emph{Advances in Neural Information
  Processing Systems}, 2019, pp. 10\,090--10\,100.

\bibitem{whitley1994genetic}
D.~Whitley, ``A genetic algorithm tutorial,'' \emph{Statistics and computing},
  vol.~4, no.~2, pp. 65--85, 1994.

\bibitem{carreira2015review}
M.~A. Carreira-Perpin{\'a}n, ``A review of mean-shift algorithms for
  clustering,'' \emph{arXiv preprint arXiv:1503.00687}, 2015.

\bibitem{gmaps}
``Directions from 95 cleveland st, darlington nsw 2008, australia to 135
  cleveland st, darlington nsw 2008, australia. google maps,'' 6 October 2020,
  maps.google.com.

\bibitem{gorelick2017google}
\BIBentryALTinterwordspacing
N.~Gorelick, M.~Hancher, M.~Dixon, S.~Ilyushchenko, D.~Thau, and R.~Moore,
  ``Google earth engine: Planetary-scale geospatial analysis for everyone,''
  \emph{Remote Sensing of Environment}, 2017. [Online]. Available:
  \url{https://doi.org/10.1016/j.rse.2017.06.031}
\BIBentrySTDinterwordspacing

\end{thebibliography}

\end{document}